\documentclass{article}

\usepackage{microtype}
\usepackage{graphicx}
\usepackage{subfigure}
\usepackage{booktabs} 

\usepackage{hyperref}

\usepackage[accepted]{icml2024}

\usepackage{amsmath}
\usepackage{amssymb}
\usepackage{mathtools}
\usepackage{amsthm}
\usepackage{caption}
\usepackage[capitalize,noabbrev]{cleveref}

\theoremstyle{plain}

\theoremstyle{definition}
\newtheorem{definition}{Definition}
\newtheorem{assumption}{Assumption}
\theoremstyle{remark}

\usepackage[textsize=tiny]{todonotes}

\icmltitlerunning{}

\begin{document}

\twocolumn[
\icmltitle{Robust Conformal Prediction under Distribution Shift via \\Physics-Informed Structural Causal Model}



\icmlsetsymbol{equal}{*}

\begin{icmlauthorlist}
\icmlauthor{Rui Xu}{gz}
\icmlauthor{Yue Sun}{lu}
\icmlauthor{Chao Chen}{lu}
\icmlauthor{Parv Venkitasubramaniam}{lu}
\icmlauthor{Sihong Xie}{gz}
\end{icmlauthorlist}

\icmlaffiliation{gz}{Information Hub AI Thrust, The Hong Kong University of Science and Technology (Guangzhou), Guangzhou, China}
\icmlaffiliation{lu}{Department of Electrical and Computer Engineering, Lehigh University, Bethlehem, USA}
\icmlcorrespondingauthor{Sihong Xie}{sihongxie@hkust-gz.edu.cn}


\icmlkeywords{Machine Learning, ICML}

\vskip 0.3in
]



\printAffiliationsAndNotice{} 

\begin{abstract}

Uncertainty is critical to reliable decision-making with machine learning. Conformal prediction (CP) handles uncertainty by predicting a set on a test input, hoping the set to cover the true label with at least $(1-\alpha)$ confidence. This coverage can be guaranteed on test data even if the marginal distributions $P_X$ are different between calibration and test datasets. However, as it is common in practice, when the conditional distribution $P_{Y|X}$ is different on calibration and test data, the coverage is not guaranteed and it is essential to measure and minimize the coverage loss under distributional shift at \textit{all} possible confidence levels. To address these issues, we upper bound the coverage difference at all levels using the cumulative density functions of calibration and test conformal scores and Wasserstein distance. Inspired by the invariance of physics across data distributions, we propose a physics-informed structural causal model (PI-SCM) to reduce the upper bound. We validated that PI-SCM can improve coverage robustness along confidence level and test domain on a traffic speed prediction task and an epidemic spread task with multiple real-world datasets.
\end{abstract}

\section{Introduction}
The prediction accuracy of machine learning has been improved a lot by the increasing amount of available data, more powerful computation hardware, and more sophisticated algorithms. However, due to ubiquitous noises or unobservability,
prediction uncertainty remains a concern when applying machine learning to domains, such as Fintech and AI+healthcare, where stakeholders need to make critical  decisions~\cite{ryu2020sustainable,seoni2023application}. 

Conformal Prediction (CP) is a generalized framework that provides a prediction set for each test input with user-specified $(1-\alpha)$ confidence level~\cite{vovk2005algorithmic}. 
Specifically, using a trained model $f$, CP calculates conformal scores (essentially residuals between predicted and true target value) of the calibration instances, and computes the $(1-\alpha)$ quantile $V_q$ of the conformal scores. The prediction set of test input $x$ includes potential labels $\hat{y}$ with a conformal score less than $V_q$.
Under the assumption that calibration and test data share the same joint distributions of the data, a test label has at least $(1-\alpha)$ \textbf{coverage} or probability of falling into the prediction set estimated using the calibration set~\cite{shafer2008tutorial}.

In reality, the calibration and test distributions are rarely the same, and researchers have developed sophisticated techniques to address the difference in marginal distribution $P_X$ of input features $X$~\cite{xu2021conformal, ghosh2023probabilistically}. Among the methods, importance weighting \cite{tibshirani2019conformal} can ensure $(1-\alpha)$ coverage of prediction sets if the conditional distributions $P_{Y|X}$ remain the same.
However, when $P_{Y|X}$ also differs across the calibration and test data, the probability that the true test label is included in the prediction set, called \textbf{exact coverage}, is no longer $(1-\alpha)$.
It is unknown how to upper-bound the difference between $(1-\alpha)$ and the exact coverage using the divergence of two different $P_{Y|X}$, and it is also challenging to reduce the gap for all confidence levels.

\begin{figure*}[h]
\centering
\captionsetup{singlelinecheck = false, justification=justified}

  \includegraphics[scale=0.37]{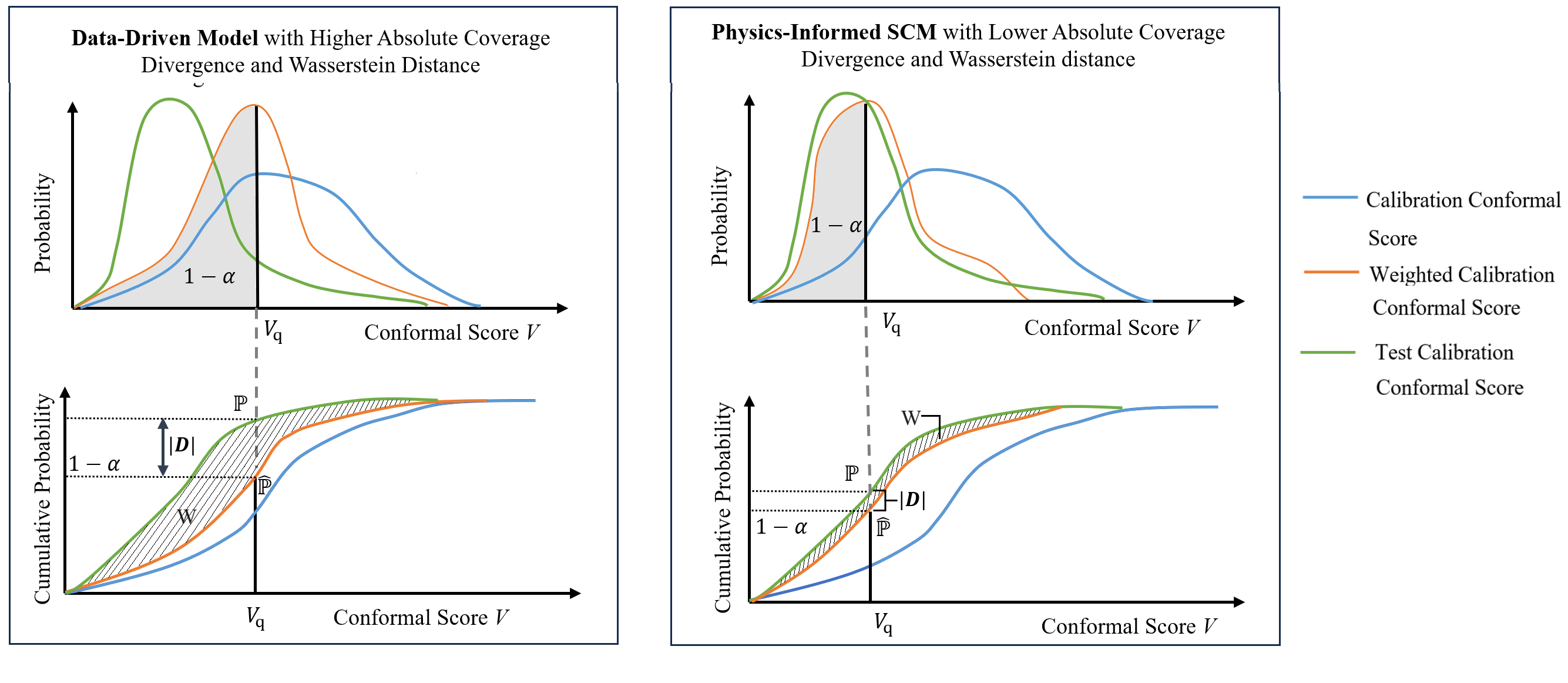}
  \vspace{-20pt}
  \caption{\textbf{Reduction of Wasserstein distance by Physics-Informed Structural Causal Model (PI-SCM).}  After reducing the influence of different marginal distributions by importance weighting, the ($1-\alpha$) quantile of weighted calibration conformal scores is calculated as $V_q$. The difference, $|D|$, between the coverage on weighted calibration conformal scores, $\hat{\mathbb{P}}$, and the coverage on test conformal scores,  $\mathbb{P}$, is calculated by their corresponding cumulative density function(CDF) at $V_q$. To evaluate the closeness of the CDFs along different confidence levels, Wasserstein distance scans $|D|$ along the quantile axis, showing the domain adaptation ability of a model. PI-SCM can capture more physical causality than data-driven models, thus leading to lower Wasserstein distance. }
  \label{fig: Coverage_Divergence} 
  \vspace{-10pt}
\end{figure*}

To address the challenges,
we first theoretically quantify coverage difference and calculate its upper bound based on the divergence between the two cumulative density functions (CDF) of calibration and test conformal scores.
To account for the upper-bound at all confidence levels,
we propose to use Wasserstein distance to integrate the upper-bound over all confidence levels to provide a comprehensive evaluation.
A small Wasserstein distance implies that the conformal predictor can generalize from calibration distribution to test distribution and thus has a low \textbf{domain adaptation generalization errors} of the conformal predictor.
The intuition of the above mechanism is illustrated in Fig.\ref{fig: Coverage_Divergence}.

Inspired by~\cite{peters2017elements}, we propose a physics-informed structural causal model (PI-SCM) to improve a model‘s domain adaptation ability by capturing the physical causality between variables within it, thus obtaining a small Wasserstein distance. Built upon these theoretical works, we proved PI-SCM can introduce more causality and thus reduce coverage divergence when test domain shifts, which is validated by experiments on a traffic speed prediction task with PeMSD4, PeMSD8, Seattle-loop datasets~\cite{guo2019attention,cui2019traffic} and an epidemic spread prediction task with US-Regions, US-States, Japan-Prefectures datasets~\cite{deng2020cola}.
\section{Background}
\subsection{Conformal Prediction}
The fundamental idea of CP is illustrated in the work~\cite{vovk2005algorithmic} and further formalized in~\cite{shafer2008tutorial}. We introduce split CP with a regression problem for ease of understanding. Suppose we have a trained model $f$, some calibration samples $(X_i^c, Y_i^c)$, $i=1...n$ included in $\textbf{S}_c$, and some test samples $(X_i^t, Y_i^t)$, $i=1...m$ included in $\textbf{S}_t$. CP introduces conformal score $V$ as the measurement of the fitness between the trained model $f$ and the calibration data. For the regression problem here, we state conformal score as the residuals between predicted calibration labels $f(X_i^c)$, and true labels $Y_i^c$. 
\begin{equation}\label{residual}
    V_i^c=|f(X_i^c)-Y_i^c|.
\end{equation}
The goal of CP is to build a prediction interval of $X_i^t$ from $\textbf{S}_t$  hoping the interval to cover the true label $Y_i^t$ with at least $(1 -\alpha)$ confidence. 

Most CP methods are based on the exchangeability assumption~\cite{vovk2005algorithmic}.
\begin{assumption}[Exchangeability]
\label{exchangeability}
$\textbf{S}_c\cup\{(X_i^t, Y_i^t)\}$ are exchangeable for any  $(X_i^t, Y_i^t) \in \textbf{S}_t$. If we name the underlying probability distribution as $P_{XY}$, then calibration data and test data share the same distribution.
\begin{equation*}
    (X_i^c, Y_i^c) \stackrel{\text{i.i.d.}}{\sim} P_{XY}\text{ , }i=1,...,n,
\end{equation*}
\begin{equation*}
    (X_i^t, Y_i^t) \stackrel{\text{i.i.d.}}{\sim} P_{XY}\text{ , }i=1,...,m.
\end{equation*}
\end{assumption}
Through Assumption~\ref{exchangeability} and Eq.~(\ref{residual}), we can state the conformal scores of calibration data and quantile data are also exchangeable with the distribution of $P_V$. $V_i^c \stackrel{\text{i.i.d}}{\sim} P_{V}$ and $V_i^t \stackrel{\text{i.i.d}}{\sim} P_{V}$. Sets of $V_i^c, i=1...n$ and $V_i^t, i=1...m$ are denoted as $\textbf{V}_c$ and $\textbf{V}_t$ respectively. Suppose a sample $(x,y) \in \textbf{S}_t$, with only $x$ is known, to obtain the prediction interval $(x, y)$ with given $\alpha$, we calculate the probability that $y \in C(x)$. $C(x)$ is the prediction interval (or prediction set for classification problems) of $x$. Suppose $\hat{y}$ is the potential true label of $x$, so $\hat{v}=|f(x)-\hat{y}|$. With the definition of quantile in $C(x)$ can be defined as:
\begin{equation}\label{CP prediction set}
    C(x)=\{\hat{y}\in \mathbb{R}:\hat{v}\leq\text{Quantile}(1-\alpha,\textbf{V}_c \cup \{\hat{v}\})\}.
\end{equation}
Eq.~(\ref{CP prediction set}) states that all $\hat{y}$ values whose conformal score is less or equal to the $(1-\alpha)$ quantile should be included in $C(x)$. It is easy to state that $y \in C(x)$ is equivalent to 
\begin{equation}\label{equivalent}
\hat{v}\leq\text{Quantile}(1-\alpha,\textbf{V}_c \cup \{\hat{v}\}).  
\end{equation}
For the explanation of quantile, we refer to Definition \ref{quantile definition}.
\begin{definition}[Quantile]\label{quantile definition}
$\text{Quantile}(1-\alpha,F)$ is denoted as the $(1-\alpha)$ quantile of distribution $Z\sim F$. 
\begin{equation}
    \text{Quantile}(1-\alpha,F)=\inf (z: \mathbb{P}(Z<z)\geq 1-\alpha). \label{quantile}
\end{equation}
With samples $(Z_1,...,Z_n)$ from $Z\sim F$ available, an empirical form is listed below, where $\delta_{Z_i}$ denoting the point mass at $Z_i$.
\begin{equation}
\text{Quantile}(1-\alpha,\frac{1}{n}\sum_{i=1}^n\delta_{Z_i}).
\end{equation}
\end{definition}
R.H.S of Eq.~(\ref{equivalent}) is intuitively saying that $\hat{v}$ is among the $\lceil 1-\alpha \rceil$ smallest members of $\textbf{V}_c \cup \{\hat{v}\}$. Due to Assumption~\ref{exchangeability} inferring $V_i^c$ and $V_i^t$ are also i.i.d. from $P_V$, the R.H.S event happens at least$\lceil \frac{(1-\alpha)(n+1)}{n+1} \rceil \geq (1-\alpha)$, so we can guarantee the \textbf{exact coverage}, $\mathbb{P}$, on test data of the prediction interval as below.
\begin{equation}\label{coverage guarantee}
    \mathbb{P}(y \in C(x))\geq (1-\alpha).
\end{equation}
We restate the R.H.S of Eq.~(\ref{equivalent}) below for the ease of assigning mass point of $\hat{v}$. To prove the equivalency, imagine a series of numbers $a_i,i=1,..,k$. After calculating $(1-\alpha)$ quantile as $q$, reassign a sample $a_i\leq q$ a value strictly larger than $q$. This modification will raise $q$ to $q'$, so $a_i\leq q'$. The possible equality still holds under the condition of $\alpha=1$ and initial $a_i=q$. 
\begin{equation}\label{infty Intro}
    \hat{v}\leq\text{Quantile}(1-\alpha,\textbf{V}_c \cup \{\infty\}).
\end{equation}
Due to the property of being distribution-free, CP attracted much attention and developed a broad family. With the assumption of the exchangeability of calibration data and test data, a lot of work focused on improving the adaptiveness of CP, such as conformalized quantile regression~\cite{Romano2019ConformalizedQR}, and classification with valid coverage~\cite{Romano2020ClassificationWV}. Researchers are also interested in improving the reliability of CP interval~\cite{ndiaye2022stable}, like cross-prediction~\cite{Vovk2012CrossconformalP} and jackknife+~\cite{Barber2019PredictiveIW}. We suggest~\cite{Zeni2020ConformalPA} to readers as a comprehensive review of CP. 
\subsection{Conformal Prediction under Covariant Shift}
The exchangeability assumption limits the application of CP as it does not always hold in reality. Importance weighting, proposed by~\cite{tibshirani2019conformal}, relaxed the Assumption~\ref{exchangeability} to conditional exchangeability as illustrated below~\cite{tibshirani2019conformal, lindley1981role}.
\begin{assumption}[Conditional Exchangeability]
\label{conditional exchangeability}
The global exchangeability among $\textbf{S}_c\cup\{(X_i^t, Y_i^t)\}$ no longer holds, but the consistency of label conditional probability distribution is preserved between calibration data and test data.
    \begin{equation*}
        (X_i^c, Y_i^c) \stackrel{\text{i.i.d.}}{\sim} P_{XY}=P_X\times P_{Y|X}, \; i=1,...,n,
    \end{equation*}
    \begin{equation*}
        (X_i^t, Y_i^t) \stackrel{\text{i.i.d.}}{\sim} \tilde P_{XY}=\tilde P_X\times  P_{Y|X}, \;i=1,...,m.
    \end{equation*}\label{conditional exchangeable test data}
    \vspace{-20pt}
\end{assumption}
Conformal scores are weighted by the likelihood ratio $w(X_i)=\frac{d\tilde P_X(X_i)}{dP_X(X_i)}$. For a sample $(x,y)$ from test data, the weighted empirical distribution $F(x)$ is listed as follows, where $p_i^c$ and $p_i^t$ are weights of point mass $\delta _{V_i^c}$ and $\delta _{\infty}$.
\begin{equation}\label{weighted cal score}
    F(x)=\sum_{i=1}^n p_i^c(x)\delta _{V_i^c}+p^t(x)\delta _{\infty},
\end{equation}
\begin{equation}\label{cal weight}
    p_i^c(x) = \frac{w(X_i^c)}{\sum_{j=1}^n w(X_j^c) +w(x)},
\end{equation}
\begin{equation}\label{test weight}
    p^t(x) = \frac{w(x)}{\sum_{j=1}^n w(X_j^c) +w(x)}.
\end{equation}
$F(x)$ is a function of $x$ but actually, this dependence will become inferior with $n>>1$. The Eq.~(\ref{CP prediction set}) is revised by the weighted conformal score distribution in Eq.~(\ref{weighted cal score}) as below, and the coverage guarantee with Eq.~(\ref{coverage guarantee})  still holds.
\begin{equation}\label{weighted coverage guarantee}
    C(x)=\{\hat{y}\in \mathbb{R}:\hat{v}\leq\text{Quantile}(1-\alpha,F(x))\}.
\end{equation}
\cite{barber2023conformal} further employed weighted quantile against distribution shift and applied it to nonsymmetric algorithms. Meanwhile,~\cite{xu2021conformal} focus on making CP more suitable for dynamic data distributions, which are likely to occur on time-series data, by aggregating bootstrap estimators to calculate prediction intervals based on the recent new samples, and~\cite{gibbs2021adaptive} approximate historical miscoverage frequency to adjust target coverage level in the next time step. All these works above successfully improved the feasibility of CP under specific interventions, but we are more interested in providing a generalized framework to address the condition that Eq.~(\ref{conditional exchangeable test data}) does not hold.
\section{Methodology}
\subsection{Coverage Divergence under Non-exchangeability}
We focus on quantifying the difference between the coverage on test data and $(1-\alpha)$, as neither Assumption~\ref{exchangeability} nor Assumption~\ref{conditional exchangeability} can be satisfied in real-world scenarios perfectly, even if the latter is relaxed. We state the condition of non-exchangeability against Assumption~\ref{exchangeability} and~\ref{conditional exchangeability}.
\begin{assumption}[Non-exchangeability]
There is no probability distribution consistency between calibration data $\textbf{S}_c$ and test data $\textbf{S}_t$.
    \begin{equation*}
        (X_i^c, Y_i^c) \stackrel{\textnormal{i.i.d.}}{\sim} P_{XY}=P_X\times P_{Y|X}, \; i=1,...,n,
    \end{equation*}
    \begin{equation*}
        (X_i^t, Y_i^t) \stackrel{\textnormal{i.i.d.}}{\sim} \tilde P_{XY}=\tilde P_X\times  \tilde P_{Y|X}, \; i=1,...,m.
    \end{equation*}
\end{assumption}
The distribution shift between $P_X$ and $\tilde P_X$ can be corrected by the likelihood ratio, $w(X_i)=\frac{d\tilde P_X(X_i)}{dP_X(X_i)}$, so we inherit the result from Eq.~(\ref{weighted coverage guarantee}). Denote $V_q$ as the $1-\alpha$ quantile of $F(x)$ as below. We can see $V_q(1-\alpha,x)$ is a function of $\alpha$ and $x$, and it is written as $V_q$ for brevity.
\begin{equation}\label{weighted quantile}
    V_q(1-\alpha,x) = \text{Quantile}(1-\alpha,F(x)).
\end{equation}
The guaranteed coverage $(1-\alpha)$ by importance weighting is, in fact, the probability that the value of $\hat{v}$ falls in the $(1-\alpha)$ quantile of $F(x)$. However, in the non-exchangeability condition, this coverage guarantee only works for weighted calibration conformal scores, but not for test data conformal scores. $\hat{\mathbb{P}}$ denotes the \textbf{expected coverage} on weighted calibration data.
\begin{equation}\label{ideal coverage}
    \hat{\mathbb{P}}(y \in C(x))=\left[\sum_{V_i^c\leq V_q}^n p_i^c(x)\delta _{V_i^c}\right]\geq (1-\alpha).
\end{equation}
Notice in Eq.~(\ref{ideal coverage}), we do not consider the trivial condition that $\alpha=1$, so $V_q<\infty$ and $\delta_\infty$ are not included there. Meanwhile, Eq.~(\ref{coverage guarantee}) does not hold, and the \textbf{exact coverage} of $C(x)$, denoted by $\mathbb{P}$, on test data changes to 
\begin{equation}\label{real coverage}
    \mathbb{P}(y \in C(x))=\frac{1}{m}\sum_{V_i^t\leq V_q}\delta_{V_i^t}.
\end{equation}
Although the true values of $\textbf{V}_t$ are unknown, we here use them as a standard to evaluate how far the coverage divergence would be. Based on Eq.~(\ref{ideal coverage}) and Eq.~(\ref{real coverage}), we can provide the upper bound of the difference between exact coverage $\mathbb{P}$ and user-specified confidence level.
\begin{equation}\label{inequality}
    \sum_{V_i^c\leq V_q}^n p_i^c(x)\delta _{V_i^c}-\frac{1}{m}\sum_{V_i^t\leq V_q}\delta_{V_i^t}\geq(1-\alpha)-\mathbb{P}(y \in C(x)).
\end{equation}
The L.H.S. of Eq.~(\ref{inequality}) quantifies the difference between the exact coverage $\mathbb{P}$ and the expected coverage $\hat{\mathbb{P}}$. We call it \textbf{coverage divergence} $D$ as 
\begin{equation}
    D(V_q) = \sum_{V_i^c\leq V_q}^n p_i^c(x)\delta _{V_i^c}-\frac{1}{m}\sum_{V_i^t\leq V_q}\delta_{V_i^t}.
\end{equation}

As we want to minimize the difference between $\mathbb{P}(y \in C(x))$ and $1-\alpha$, so we should minimize the \textbf{absolute coverage divergence} $|D(V_q)|$.

$|D(V_q)|$ measures the vertical distance between the empirical CDFs of test conformal scores and weighted calibration conformal scores via importance weighting. The relationship is illustrated in Fig.\ref{fig: Coverage_Divergence}. Since $|D(V_q)|$ is a function of $V_q$, and the value of $V_q$ is related to $\alpha$, $|D(V_q)|$ will fluctuate along $\alpha$ as well. Nevertheless, we hope a model with good \textbf{coverage robustness}, i.e. keeping small $|D|$ with different confidence levels and different test sets, is preferred. 

To build a comprehensive evaluation of how far the exact coverage would diverge from expected coverage, we apply the concept of \textbf{Wasserstein distance} as an integrator of $|D(V_q)|$ along all quantile values, which is in fact along all $(1-\alpha)$ confidence levels
\begin{equation}\label{W distance}
    W=\sum_{V_q\in \mathbb R}|D(V_q)|.
\end{equation}
The core reason why $W$ exists is the inconsistency of label conditional distribution between test data and calibration data, and this inconsistency is reflected in the conditional distribution of conformal scores. Thus, it is necessary to obtain a model that has small domain generalization errors by accurately capturing the causality between features and labels. 
Instead, applying pure data-driven models is likely to lead to overfitting.\\

\subsection{Physics-Informed Structural Causality Model}

\textbf{Structural Causality Model (SCM)} is applied as it can be a framework for formulating a model with consistent conditional distribution when the data domain shifts (see e.g.,~\cite{pearl2009causality}). Assume the data generation process follows a SCM, $\mathcal{M}$, having system variables $\textbf{A}_{\mathcal{J}}=\{A_j\}_{j\in\mathcal{J}}$, exogenous unobserved independent  variables $\textbf{E}_{\mathcal{K}}=\{E_k\}_{k\in\mathcal{K}}$, and context variables $\textbf{U}_{\mathcal{I}}=\{U_i\}_{i\in\mathcal{I}}$. $PA(\cdot)$ indicates the parent of a variable, i.e. directly caused by $PA(\cdot)$.

The SCM, $\mathcal{M}$, can be written as a causal graph $\mathcal{G}(\mathcal{M})$. Here we use implicit representation, indicating the presence of noise is implied by the fact that the structural equations contain error terms. So, we do not draw noise variables explicitly and $\mathcal{G}(\mathcal{M})$ only contains system variable nodes and context variable nodes, denoted by $\mathcal{I}\cup\mathcal{J}$. In $\mathcal{G}(\mathcal{M})$, directed edges $n_1\rightarrow n_2$ for $n_1,n_2 \in \mathcal{I}\cup\mathcal{J}$ iff $n_1\in PA(n_2)$, and bidirected edges $n_1\leftrightarrow n_2$ iff the existence of $k\in PA(n_1)\cap PA(n_2)\cap \mathcal{K}$, indicating unmeasured variables $E_k$ that affect $n_1$ and $n_2$ within the causal graph.

Based on this notation, the SCM should obey two crucial Joint Causal Inference assumptions to clarify the relationship between system variables and context variables~\cite{mooij2020joint}. \\

\begin{assumption}[Joint Causal Inference]
$\mathcal{G}(\mathcal{M})$ is a causal graph with system variables $\{A_j\}_{j\in\mathcal{J}}$ and context variables $\{U_i\}_{i\in\mathcal{I}}$.\\
1. Exogeneity Assumption: Any system variables do not directly cause any context variables.\\
\centerline{$\forall i\in \mathcal{I}, j \in \mathcal{J}: A_j\rightarrow U_i\notin \mathcal{G}(\mathcal{M})$}
2. Randomization Assumption: Any system variable is not confounded with a context variable.\\
\centerline{$\forall i\in \mathcal{I}, j \in \mathcal{J}: A_j\leftrightarrow U_i\notin \mathcal{G}(\mathcal{M})$}
\end{assumption}
\newpage
With these assumptions, we can write the representation of the model $\mathcal{M}$ as Eq.~(\ref{SCM}). 
\begin{equation}\label{SCM}
    \begin{cases}
    U_i=g_i(\textbf{E}_{PA(i)\cap\mathcal{K}}), & i\in \mathcal{I}\\
    A_j=f_j(\textbf{A}_{PA(j)\cap\mathcal{J}},\textbf{U}_{PA(j)\cap\mathcal{I}},\textbf{E}_{PA(j)\cap\mathcal{K}}),&j\in \mathcal{J}\\
    \mathbb{P}(\textbf{E})=\prod_{k\in\mathcal{K}} \mathbb{P}(E_k).
        
    \end{cases}
\end{equation}
This formula shows that $\{U_i\}_{i\in\mathcal{I}}$ is independent from $\{A_j\}_{j\in\mathcal{J}}$, whereas $\{A_j\}_{j\in\mathcal{J}}$ are still having causality from $\{U_i\}_{i\in\mathcal{I}}$.

If $B=A_1\in\{A_j\}_{j\in\mathcal{J}}$ is the label we are interested in, whose parent is $\mathcal{P}^B=\textbf{A}_{PA(1)\cap\mathcal{J}}\cup\textbf{U}_{PA(1)\cap\mathcal{I}}\cup\textbf{E}_{PA(1)\cap\mathcal{K}}$,
and $U_1\notin \mathcal{P}^B$ is the context variable indicating the training domain or test domain by 0 or 1, the data generation mechanism of the SCM should be consistent no matter how the value $U_1$ changes, i.e. $U_1\perp B |\mathcal{P}^B[\mathcal{G}(\mathcal{M})]$. This consistency can be written like Assumption~\ref{conditional exchangeability} as the following to introduce more physical causality and obtain better domain adaptation ability.
\begin{equation}\label{SCM conditional probability}
    P(B|\mathcal{P}^B, U_1=1)=P(B|\mathcal{P}^B, U_1=0).
\end{equation}
However, the true relationship $f_1$ between $B$ and its parent $\mathcal{P}^B$ is unknown. To build a predictor $\hat{f}_1$ that has a good domain adaptation ability, it is necessary to capture the physical causality underneath the data.  

\textbf{Physics-Informed Structural Causal Model (PI-SCM)} is proposed to recognize the physical relationships.
Imagein we train a predictor $\hat{f_1}$ of $B$ based on observed parent $\mathcal{P}^B_{ob}=\textbf{A}_{PA(1)\cap\mathcal{J}}\cup\textbf{U}_{PA(1)\cap\mathcal{I}}$, to minimize the loss $\mathbb{E}((\hat{B}-B)^2|U_1=0)$ on training domain, where the predicted value is $\hat{B}=\hat{f}(\mathcal{P}^B_{ob})$. If $\hat{f_1}$ is physics-guided with perfect causality, the conditional probability of residual $R=(\hat{B}-B)^2$ should be the same.
\begin{equation}\label{perfect SCM}
    P(R|\mathcal{P}^B_{ob}, U_1=1)=P(R|\mathcal{P}^B_{ob}, U_1=0).
\end{equation}
Eq.~(\ref{perfect SCM}) indicates two properties of PI-SCM. First, the observed variables are enough to guarantee the domain adaptation ability of $\hat{f}_1$, and unmeasured exogenous variables do not result in biased estimation. Secondly, $\hat{f}_1$ is fully physics-informed instead of a data-driven model, so it will not overfit partial data and cause outliers of residuals, which will break the equality in Eq.~(\ref{perfect SCM}).

However, in reality, Eq.~(\ref{perfect SCM}) may not always hold due to the lack of available data and insufficient knowledge of the physics relationship, so we can hardly eliminate the domain generalization error from latent variables $\{E_k\}_{k\in\mathcal{K}}$. Because of that, it is necessary to measure how big the difference of $P(R|\mathcal{P}^B_{ob})$ is under the context of $U_1=1$ and $0$, and the Wasserstein distance introduced by Eq.~(\ref{W distance}) can act as a gauge to quantify the closeness between a selected model and PI-SCM.

\section{Experiment}
We conducted experiments on two tasks: traffic speed prediction and epidemic spread prediction, implemented via~\cite{scikit-learn}. In each case, we demonstrate that a model with more causality will lead to more robust desired coverages.  
\subsection{Traffic Speed Prediction}
\subsubsection{Reaction Diffusion Model}
Reaction-diffusion (RD) mechanism is introduced to traffic systems for speed prediction to discover the underlying traffic patterns of different road segments~\cite{bellocchi2020unraveling,sun2023reaction}. Intuitively, for the traffic speed $u_i(t)$ at segment $i$ and moment $t$, the reaction term and diffusion term explain the influence from $N^{r}$ downstream and $N^{d}$ upstream road segments respectively. RD-U model only considers the interactions in terms of speed.
\begin{align}
\triangle u_i(t)&=\sum_{j\in N^d}\rho_{(i,j)}\triangle u_{(i,j)}(t)+d_i \notag\\ &+\tanh({\sum_{j\in N^r}\sigma_{(i,j)}\triangle u_{(i,j)}(t)+r_i})\label{RD-U}.
\end{align}
where $\triangle u_i(t) = u_i(t+\delta t) - u_i(t)$ with time step $\delta t$, and $\triangle u_{(i,j)}(t) = u_j(t)-u_i(t)$, i.e. the speed difference of two locations. $\rho_{(i,j)}$ and $\sigma_{(i,j)}$ are diffusion and reaction parameters for segment $i$ and $j$. 

Here we propose the RD-UQ model as below, including traffic volume $q$, and conduct data separation by traffic density. These actions will include more causality to the RD-UQ model.  For a detailed causality analysis of the RD-UQ model, we refer to the Appendix.\ref{A:RD}. 
After defining $\triangle q_{(i,j)}(t) = q_j(t)-q_i(t)$, RD-UQ model can be written as below. $\rho_{(i,j)}^u$, $\rho_{(i,j)}^q$, $\sigma_{(i,j)}^u$, $\sigma_{(i,j)}^q$ are diffusion and reaction parameters of speed and traffic volume between segment $i$ and $j$.
\begin{align}
&\triangle u_i(t)=\sum_{j\in N^d}(\rho_{(i,j)}^u\triangle u_{(i,j)}(t)+\rho_{(i,j)}^q\triangle q_{(i,j)}(t))+d_i \notag\\ &+\tanh(\sum_{j\in N^r}\sigma_{(i,j)}^u\triangle u_{(i,j)}(t)+\sigma_{(i,j)}^q\triangle q_{(i,j)}(t)+r_i).\label{RD-QU}
\end{align}
\subsubsection{Experiment Setup}
We applied the RD-U and RD-UQ models on Seattle-loop~\cite{cui2019traffic}, and PeMSD4, PeMSED8 datasets~\cite{guo2019attention}. The time interval $\delta 
t$ between sensor snapshots is 5 minutes for the three datasets. For the ease of kernel density estimation for probability distributions, we only select nodes with the degree of 2 for the experiment, so each sensor is only connected to a single upstream neighbor and a downstream neighbor. The experiment is conducted on workday data.  Training and calibration phases are operated based on the 24-hour measurement. Test data is further separated into single-hour slots to evaluate the model's coverage robustness with different test distributions. So, for each segment, we will have one training set, one calibration set, and 24 test sets. $\alpha$ ranges from 0.1 to 0.9 by 0.1 step size.

\textbf{Seattle-loop}: The dataset contains the traffic data of the Seattle area in 2015.  The freeways contain I-5, I-405, I-90, and SR-520, and we select 61 2-degree nodes from them. For each node, the ratio of training data, calibration data, and test data is 35$\%$(91 days): 15$\%$(39 days): 50$\%$(130 days). \\
\textbf{PeMSD4}: It refers to the traffic data of 29 roads in San Francisco from January to February 2018. 85 nodes are randomly selected.  The ratio of training data, calibration data, and test data is 50$\%$(21 days): 25$\%$(11 days): 25$\%$(11 days).\\
\textbf{PeMSD8}: It includes the traffic data of 8 roads in San Bernardino from July to August 2016. 33 nodes are randomly selected. The ratio of training data, calibration data, and test data is 50$\%$(22 days): 25$\%$(11 days): 25$\%$(11 days).

Kernel density estimation (KDE) is applied when calculating the value of $w(X_i)=\frac{d\tilde P_X(X_i)}{dP_X(X_i)}$. To conduct a fair comparison between the RD-U model and the RD-UQ model, it is necessary to choose the best bandwidth for each of them. As Fig.\ref{fig: bandwidth_impact}shows, we tried multiple choices of bandwidth values and applied grid search by~\cite{scikit-learn} and proved the bandwidth optimized by grid search leads to the smallest coverage divergence $|D|_{\alpha}$.We refer to the Appendix.\ref{C: BS} for the optimization of bandwidth work.

\subsection{Epidemic Spread Prediction}
\subsubsection{SIR Model and SIS Model}
To further validate the effectiveness of PI-SCM, we compare the coverage robustness of the SIR model and SIS model on the task of epidemic spread prediction. The SIR model divides the population into three categories: susceptible to the disease $S$, infectious $I$, and recovered with immunity $R$, and describes the temporal dynamic changes of their populations~\cite{cooper2020sir}. The governing differential equations can be written as
\begin{equation}
\begin{cases}
    \frac{dS(t)}{dt}=\frac{-\beta S(t)I(t)}{N},\\
    \frac{dI(t)}{dt}=\frac{\beta S(t)I(t)}{N} - \gamma I(t)=(\frac{\beta S(t)}{N} - \gamma)I(t),\\
    \frac{dR(t)}{dt}= \gamma I(t).
    \end{cases}
\end{equation}
where $N$ is the constant total population, $\beta$ is infection rate, and $\gamma$ is recovery rate. We assume the location is isolated, so $N=S(t)+I(t)+R(t)$. Also, the population of recovered people is $R(t)=\gamma \int_0^t I(t) dt$. Based on this, if $t_o$ is the starting time of the current epidemic and $\triangle I(t)$ represents the change at moment $t$, we can rewrite the dynamic change of infectious people in a discrete form as 
\begin{equation}
    \triangle I(t)=(\frac{\beta (N-I(t)-\gamma\sum_{t_o}^t I(t))}{N}-\gamma)I(t).
\end{equation}
However, the SIS model does not consider immunity from recovery and regards the recovered population as susceptible again~\cite{gray2011stochastic}. 
\begin{equation}
    \begin{cases}
        \frac{dS(t)}{dt}=\frac{-\beta S(t)I(t)+\gamma I(t)}{N},\\
        \frac{dI(t)}{dt}=(\frac{\beta S(t)}{N} - \gamma)I(t).
    \end{cases}
\end{equation}
As there is no $R$ term in the SIS model, $\triangle I(t)$ is simpler as
\begin{equation}
    \triangle I(t)=(\frac{\beta (N-I(t))}{N}-\gamma)I(t).
\end{equation}
In this experiment, we consider influenza-like illness (ILI) as the disease we want to predict. According to~\cite{deng2020cola,patel2021immune}, infection on ILI will provide temporary immunity. In this case, the SIR model fits PI-SCM better as it takes immunity as a variable in the system. This immunity is not long-lasting and may not provide protection against other strains of the influenza virus, so we reset $R(t)$ as zero at the beginning of every epidemic period, usually a year. 
\subsubsection{Experiment Setup}
We applied the SIR model and SIS model on three epidemic datasets,  US-Regions, US-States, and Japan-Prefectures~\cite{deng2020cola}.

\textbf{US-Regions}: This dataset contains the ILINet part of the U.S. Department of Health and Human Services dataset, including the weekly influenza infection count of 10 locations in the U.S. mainland from 2002 to 2017.\\
\textbf{US-States}: This is a dataset from the Center
for Disease Control and Prevention (CDC) including the number of patients infected by ILI for each week from 2010 to 2017.\\
\textbf{Japan-Prefectures}: This dataset is collected from the Infectious Diseases Weekly Report in Japan, containing weekly ILI patient counts of 47 prefectures from August 2012 to March 2019.

We divide the data into training, calibration, and test data by the ratio of $35\%:35\%:30\%$. $\alpha$ range is again $[0.1, 0.2,...,0.9]$. To obtain coverage robustness performance on different test data distributions, we further divide the test data into four pandemic intervals, Initiation, Acceleration, Declaration, and Subsidence, according to the Pandemic Intervals Framework (PIF) by CDC. Appendix.\ref{B.PI} shows the standard for separating pandemic intervals.

\section{Result}
\subsection{Evaluation Metrics}
We consider multiple aspects to evaluate models:\\ 
\textbf{Absolute Coverage Divergence}:  For each $\alpha$, we calculate absolute coverage divergences $|D|_{\alpha}$ to study the relationship between the divergence and confidence level. We also compute the averaged $|D|_{t}$ to show how the coverage robustness is improved among different test domains. \\
\textbf{Prediction Size}: We do not want to increase the prediction size too much as a sacrifice for coverage robustness, so the prediction size along $(1-\alpha)$ is also calculated.\\ 
\textbf{Prediction Accuracy}: models' accuracy in terms of root mean square error (RMSE) and mean absolute error (MAE) is measured.

\subsection{Coverage Robustness along Confidence Level}
\begin{figure}[h]
\vspace{25pt}
\captionsetup{singlelinecheck = false, justification=justified}
\centering
\includegraphics[scale=0.23]{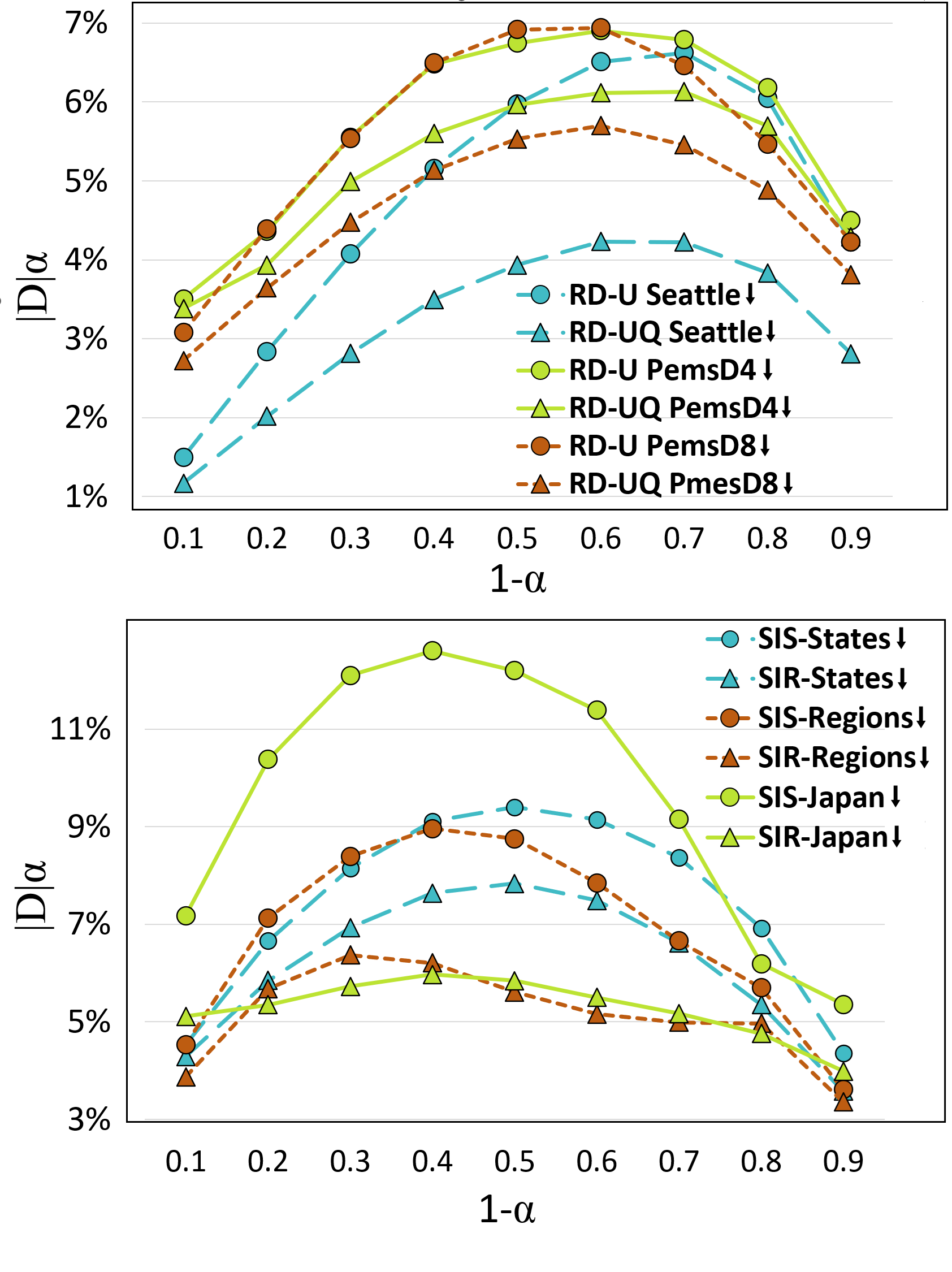}
\vspace{-9pt}
\caption{\textbf{Absolute coverage divergences $|D|_{\alpha}$ along $(1-\alpha)$ confidence level of traffic speed prediction (top) and epidemic spread prediction (bottom)}.  Models better fit PI-SCM introduce more physical causality and diverge less from expected coverage, thus showing better coverage robustness. The result is averaged over 10 runs.}
\label{fig: Overall_Divergence_result}
\vspace{-15pt}
\end{figure}

Fig.\ref{fig: Overall_Divergence_result} demonstrates PI-SCM reduces absolute coverage divergence $|D|_\alpha$ no matter how the confidence level changes by capturing physical relationships to the model. Trained predictors based on physics-informed models lead to smaller domain generalization errors as they can capture the underlying mechanisms of data generation better and act as more universal laws during prediction. Even if in the experiments we did not directly calculate the Wasserstein distance, $W$, between the CDFs of weighted calibration conformal scores and test conformal scores, proposed in Eq.~(\ref{W distance}), the corresponding area covered under each $|D|_\alpha$ curve in Fig.\ref{fig: Overall_Divergence_result} can be regarded as an approximation of $W$. Physics-informed models have lower $W$ and this result validated the coverage robustness introduced by PI-SCM as well. 

We can also observe some interesting phenomena in Fig.\ref{fig: Overall_Divergence_result}. First, $|D|_\alpha$ tends to be high around the middle of the $(1-\alpha)$ range and to be low at two endpoints, showing an increasing-decreasing pattern, no matter what model was applied. Besides, this pattern is more obvious in models with less physical causality. This observation can be explained intuitively in Fig.(\ref{fig: Coverage_Divergence}) by the light-tailed shape of conformal score distributions. Even if $V_q$, the quantile of weighted calibration conformal scores calculated by Eq.~(\ref{weighted quantile}), does not lead to $(1-\alpha)$ coverage on test sets, when $(1-\alpha)$ is very low or high, the error in quantile will not cause high $|D|_\alpha$ due to low score density there. However, when $(1-\alpha)$ is at the middle of the range,  small quantile errors can result in high $|D|_\alpha$.  That is the reason why the increasing-decreasing pattern exists and why models guided by PI-SCM have more improvement around the middle value of $(1-\alpha)$.

\begin{figure}[h]
\captionsetup{singlelinecheck = false, justification=justified}
\centering
\includegraphics[scale=0.4]{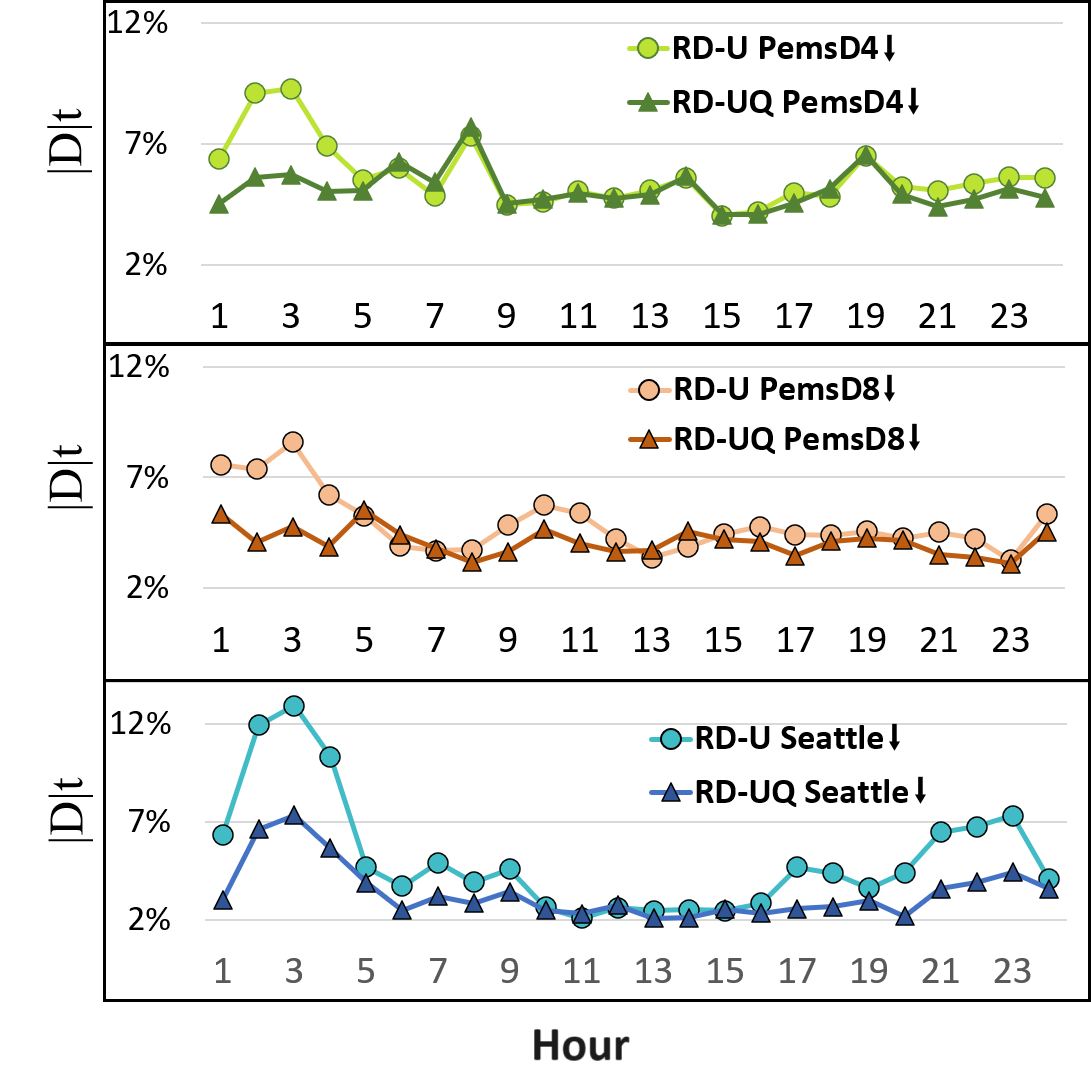}
\vspace{-5pt}
\caption{\textbf{Absolute coverage divergences $|D|_{t}$ along single-hour test sets of traffic speed prediction task}. RD-UQ model guided by PI-SCM reduces high $|D|_{t}$ of RD-U model (like from 1:00 AM to 5:00 AM) to the low level of other hours. The result is averaged over 10 runs.}
\label{fig: Traffic_Divergence_hour}

\end{figure}

\subsection{Coverage Robustness along Test Domains }
Fig.\ref{fig: Traffic_Divergence_hour} illustrates RD-UQ model guided by PI-SCM reduces high $|D|_{t}$ of the RD-U model,i.e. from 1:00 to 5:00, to the low level of other hours, indicating it overcomes the overfitting of RD-U model by capturing the physical relationship between traffic speed and volume. Meanwhile, RD-UQ model does not increase those low $|D|_{t}$ of the RD-U model, i.e. daytime hours, as a sacrifice. This phenomenon demonstrates that PI-SCM does not work as minimax algorithms focusing on reducing the highest risk of prediction. Instead, PI-SCM uses physics causality to minimize domain generalization error and is capable of providing similar low $|D|_{t}$ across all single-hour test sets.

\begin{table*}[t]
\vspace{-7pt}
\small
\centering
\captionsetup{font=small}
\captionsetup{justification=justified}
\caption{\textbf{Traffic Speed Prediction Accuracy}}
\vspace{-10pt}
\begin{tabular}{ |p{1.2cm}||p{2cm}|p{2cm}|p{2cm}|p{2cm}|p{2cm}|p{2cm}|  }
 \hline
 &\multicolumn{3}{|c|}{RMSE (average$\pm$ standard deviation)} & \multicolumn{3}{|c|}{MAE (average$\pm$ standard deviation)}\\
 \hline
 Method& Seattle-loop & PemsD4 & PemsD8& Seattle-loop & PemsD4 & PemsD8\\
 \hline
 RD-U   & 4.50 $\pm$ 2.4E-3    &1.42 $\pm$ 1.6E-3&   0.82 $\pm$ 2.3E-3 & 3.06 $\pm$ 1.5E-3 & 0.79 $\pm$ 1.9E-3 & 0.48 $\pm$ 8E-4 \\
 RD-UQ  &   \textbf{4.40 $\pm$ 2.8E-3}  & \textbf{1.41 $\pm$ 2.9E-3}   &\textbf{0.81 $\pm$ 1.7E-3} & \textbf{3.01 $\pm$ 1.4E-3} & \textbf{0.78 $\pm$ 1.0E-4} & \textbf{0.47 $\pm$ 4E-4}\\
 \hline
\end{tabular}
\label{Traffic Prediction Accuracy}
\end{table*}

\begin{figure}[h]
\captionsetup{singlelinecheck = false, justification=justified}

\centering
\includegraphics[scale=0.27]{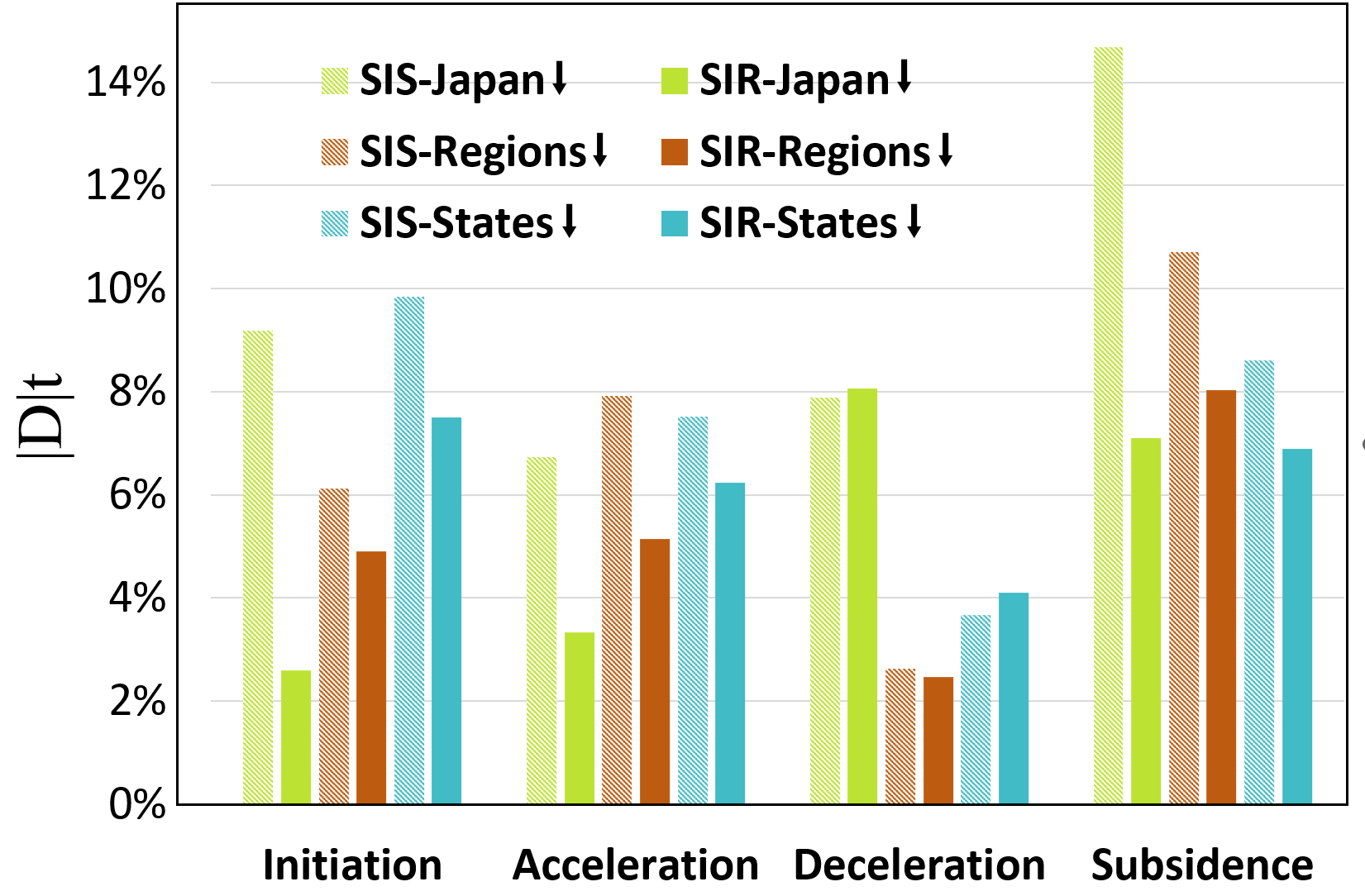}
\caption{\textbf{Absolute coverage divergences $|D|_{t}$ along pandemic intervals of epidemic speed prediction task}. SIR model reduces $|D|_{t}$ in Initiation, Acceleration, and Subsidence intervals, whereas the improvement in Deceleration interval is not obvious. The result is averaged over 10 runs.}
\label{fig: Epidemic_Divergence_Interval}
\vspace{-12pt}
\end{figure}

\begin{table*}[t]
\small
\vspace{-8pt}
\centering
\captionsetup{font=small}
\captionsetup{justification=justified}
\caption{\textbf{Epidemic Spread Prediction Accuracy}}
\vspace{-10pt}
\begin{tabular}{ |p{1.2cm}||p{1.85cm}|p{1.85cm}|p{2.3cm}|p{1.85cm}|p{1.85cm}|p{2.3cm}|  }
 \hline
 &\multicolumn{3}{|c|}{RMSE (average$\pm$ standard deviation)} & \multicolumn{3}{|c|}{MAE (average$\pm$ standard deviation)}\\
 \hline
 Method& US-Regions & US-States & Japan-Prefectures& US-Regions & US-States & Japan-Prefectures\\
 \hline
 SIS   & 366.53 $\pm$ 2.70    &110.77 $\pm$ 1.20&   656.11 $\pm$ 2.82 & 151.02 $\pm$ 0.77 & 39.83 $\pm$ 0.09 & 208.52 $\pm$ 0.77 \\
 SIR  &   \textbf{360.17 $\pm$ 3.01}  & \textbf{106.46 $\pm$ 1.00}   &\textbf{589.32 $\pm$ 3.64} & \textbf{149.94 $\pm$ 0.27} & \textbf{38.31 $\pm$ 0.11} & \textbf{178.11 $\pm$ 0.69}\\
 \hline
\end{tabular}
\label{Epidemic Prediction Accuracy}
\vspace{-10pt}
\end{table*}
Fig.\ref{fig: Epidemic_Divergence_Interval} shows SIR model provides more robust coverage than the SIS model by considering the temporal immunity of the recovered population as a system variable. However, this comparison also presents a weakness of the SIR model, indicating it is still far from a PI-SCM. No obvious reduction of $|D|_{t}$ shows up in the Deceleration interval, meaning the mechanisms of the decreasing number of reported positive-ILI patients are unclear. Multiple factors make work at this interval. For instance, immunity from vaccines can protect people from being infected, which will speed up the reduction of the reported count of patients. SIRV model, where V considers the vaccinated population, may work better in this condition~\cite{ameen2020efficient}. 

\subsection{Prediction Size and Prediction Accuracy}
Experimental results of prediction size along $(1-\alpha)$ are listed in Appendix.\ref{D: PS}. To increase the precision and usefulness of our predictions, a small prediction size is preferred under the coverage guarantee. Appendix.\ref{D: PS} shows that PI-SCM does not increase prediction size as a sacrifice of coverage robustness. Moreover, aided by PI-SCM, prediction size may be reduced at high confidence levels. This indicates physical causality can make some outliers of conformal scores closer to the main body of score distribution. 

Table.\ref{Traffic Prediction Accuracy} and Table.\ref{Epidemic Prediction Accuracy} show PI-SCM can also help improve prediction accuracy across different tasks. The result is averaged over 10 runs. The improvement is not that obvious and may give an illusion that the effectiveness of newly added variables, like traffic volume in the RD-UQ model, is marginal. However, if we see the improvement from the coverage robustness view, we can realize that PI-SCM, compared with other data-driven models, is trying to build physical causality among variables and construct universal laws underneath data. Following the logic of PI-SCM, even slight accuracy improvement can be a signal of lower domain generalization errors.

\section{Conclusion}
We propose PI-SCM to study the relationship between CP coverage robustness with non-exchangeable data and domain generalization error of trained predictors.  First, we theoretically quantify the coverage divergence on test data using the CDFs of conformal scores. Based on that, Wasserstein distance is applied as a comprehensive evaluation of domain generalization errors. Lastly, PI-SCM is proposed as a framework to improve models' domain generalization ability by capturing physical causality underneath available data. We applied PI-SCM to traffic speed prediction and epidemic spread forecast with real-world data and validated the effectiveness of PI-SCM across different fields.

\section{Impact Statement}
This paper presents work whose goal is to advance the field of Machine Learning. There are many potential societal consequences of our work, none of which we feel must be specifically highlighted here.

\nocite{langley00}
\newpage
\bibliography{Arxiv_Submission}
\bibliographystyle{icml2024}

\newpage
\appendix
\twocolumn

\section{Reaction Diffusion Model for Traffic Speed Prediction}\label{A:RD}
Reaction-diffusion (RD) mechanism was initially proposed in chemical systems to describe the dynamic state of particles. This mechanism is introduced to the traffic system by~\cite{bellocchi2020unraveling} as a location-specific model to discover the underlying traffic patterns of different road segments and to replace pure data-driven approaches, such as long-short-term memory. In~\cite{sun2023reaction}, Sun applied the reaction-diffusion model and embedded a domain equation, which consists of a reaction term and a diffusion term, within graphical neural networks to quantify the interactions of traffic states among neighbor road segments. The reaction term explains the influence opposite to the direction of the traffic flow, and the diffusion term tracks the influence along the direction of the traffic flow.  

We consider a sensor $i$ that has $N^{d}$ neighbor sensors on the upstream road segments and $N^{r}$ neighbor sensors on the downstream road segments, which means traffic states at $N^{d}$ and $N^{r}$ sensors will exert diffusion and reaction effect to sensor $i$ respectively. $u_i(t)$ denotes the speed measured by sensor $i$ at time $t$, and $\delta t$ is the time interval between sensor $i$ 's snapshots. We also claim $\triangle u_i(t) = u_i(t+\delta t) - u_i(t)$, $\triangle u_{(i,j)}(t) = u_j(t)-u_i(t)$. The reaction-diffusion differential equation in finite element form at sensor $i$ can be written as below, and we call it the RD-U model for convenience. $\rho_{(i,j)}$ and $\sigma_{(i,j)}$ are diffusion and reaction parameters for sensor $i$ and sensor $j$;  $d_i$ and $r_i$ act as the corresponding terms biases. 
\begin{align}
\triangle u_i(t)&=\sum_{j\in N^d}\rho_{(i,j)}\triangle u_{(i,j)}(t)+d_i \notag\\ &+\tanh({\sum_{j\in N^r}\sigma_{(i,j)}\triangle u_{(i,j)}(t)+r_i}).
\end{align}
The goal of the RD-U model is to find out the traffic pattern at sensor $i$ behind the data, as the parameter $\rho_{(i,j)},\sigma_{(i,j)},d_i^u,r_i^u$ are not functions of time, instead they are location-specific constants determined by local factors, like the number of lanes, road intersections, traffic lights, etc. Because of this, we wish the RD-U model should hold strong causality and output the expected 1-$\alpha$ confidence interval when the covariant shift happens to test sets. 
However, since human behavior can not be perfectly quantified by particle dynamics from chemistry, we claim two changes to the RD-U model, i.e. Eq.~(\ref{RD-U}), to improve its causality.

\begin{figure*}
\captionsetup{singlelinecheck = false, justification=justified}

  \includegraphics[scale=0.4]{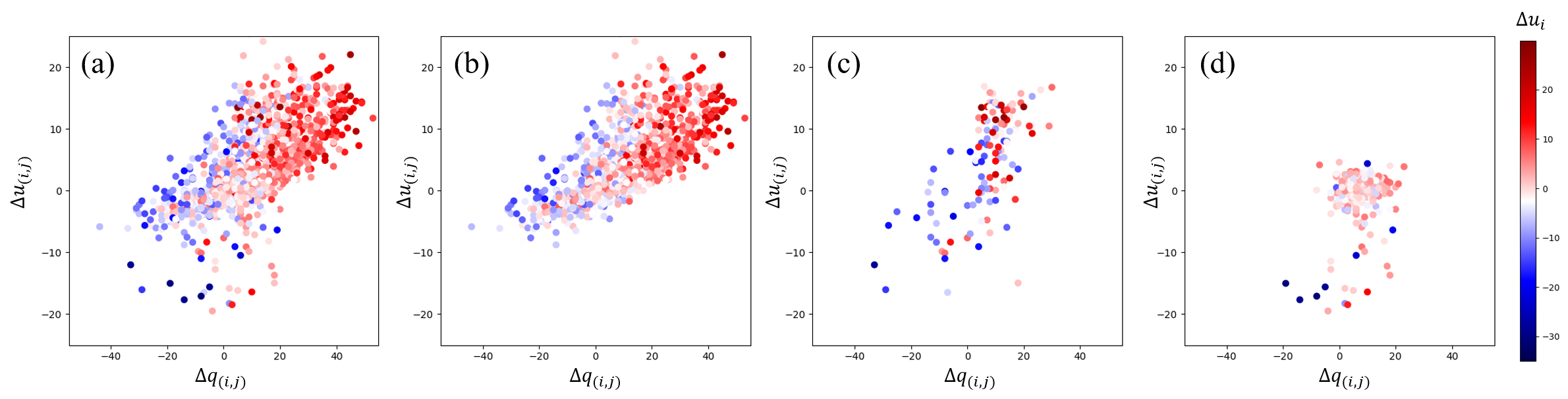}

  \caption{Reaction effect data from 7:00 AM to 8:00 AM of sensor ID: D005ES17288, whose $N^r$ has only 1 sensor for simpler presentation, in Seattle-loop dataset. (a) All $\triangle u_(i,j)$,$\triangle q_(i,j)$, and $\triangle u_i$ data of the sensor. (b) High-density traffic data, $k_i\in (k_2,\infty)$. (c) Medium-density traffic data, $k_i\in [k_1,k_2]$,(d) Low-density traffic data, $k_i\in [0,k_1)$. (a) presents $\triangle q_(i,j)$ is a strong indicator for $\triangle u_i$, whereas dependency of $\triangle u_i$ on $\triangle u_(i,j)$ is unclear without $\triangle q_(i,j)$ aid. Comparison of (b), (c), and (d) shows the correlation between $\triangle u_(i,j)$,$\triangle q_(i,j)$, and $\triangle u_i$ is stronger as traffic density,$k_i$, increases.}
  \label{fig: Traffic_Data_Investigation}
\end{figure*}

\textbf{a. Improved Causality by Traffic Volume}\\
The direct reason for traffic flow deceleration is that too many vehicles are constrained within a road segment causing a reduction in traffic volume, which measures how many vehicles pass a point, like a sensor, within a unit time interval. Only depending on the speed changes observed by neighbor sensors is not sufficient to make accurate predictions on future speed. For instance, if the downstream speed decreases whereas traffic volume increases, this phenomenon indicates the increase of vehicles there may not cause traffic congestion, so possibly will not exert a resistance reaction on local traffic flow.  Only if the downstream speed and volume decrease simultaneously does a strong signal of congestion appear there. Fig \ref{fig: Traffic_Data_Investigation}(a) shows the net effect of spatial speed and volume change can be regarded as a more reliable regressor. The diffusion effect has a similar logic. We refer to~\cite{treiber2013traffic}, which extensively illustrated the relationship between traffic volume and speed. Inspired by the existing investigations, the traffic volume information is added to the RD-U model Eq.~(\ref{RD-U}). The traffic volume difference at sensor $i$ and sensor $j$ at moment $t$ is  $\triangle q_{(i,j)}(t) = q_j(t)-q_i(t)$. The model of \ref{RD-QU} is called the RD-UQ model.
\begin{align}
&\triangle u_i(t)=\sum_{j\in N^d}(\rho_{(i,j)}^u\triangle u_{(i,j)}(t)+\rho_{(i,j)}^q\triangle q_{(i,j)}(t))+d_i \notag\\ &+\tanh(\sum_{j\in N^r}\sigma_{(i,j)}^u\triangle u_{(i,j)}(t)+\sigma_{(i,j)}^q\triangle q_{(i,j)}(t)+r_i).
\end{align}
\textbf{b. Traffic Density Informed Data Separation.}\\
The RD model tries to predict the temporal gradient of traffic speed in terms of the spatial gradient, which can be regarded as the spread of traffic waves.  However, the propagation speed of traffic waves is not constant and it is intensively correlated to traffic flow density, which is the ratio between volume $q$ and speed $u$.  We take the reaction effect as an example. Vehicles are spread out at low densities and can maneuver easily without affecting one another. Therefore, disturbances in traffic flow are less likely to cause significant traffic waves, as drivers can adjust their speed without causing a chain reaction. As the density of traffic increases, vehicles become closer together, and the freedom to maneuver decreases. At this point, small disturbances, such as a car braking or changing lanes, can have a more significant impact, causing vehicles behind to slow down.  This phenomenon is illustrated in Fig.\ref{fig: Traffic_Data_Investigation}, where we manually separate the data according to the traffic density of sensor $i$ at time $t$: $k_i(t)=q_i(t)/u_i(t)$, by two thresholds $k_1$ and $k_2$. According to the comparison of Fig.\ref{fig: Traffic_Data_Investigation} (b), (c), and (d), we can conclude the correlation between $\triangle u_(i,j)$,$\triangle q_(i,j)$, and $\triangle u_i$ is stronger as $k_i$ increases. Because of that, the RD-UQ model is trained based on the three sets separately, and we obtain three trained models in the end. The values of $k_1$ and $k_2$ are approximated in the experiment implementation.
\newpage
\section{Pandemic Intervals}\label{B.PI}
\begin{figure}[h]
\captionsetup{}
\centering
  \includegraphics[scale=0.3]{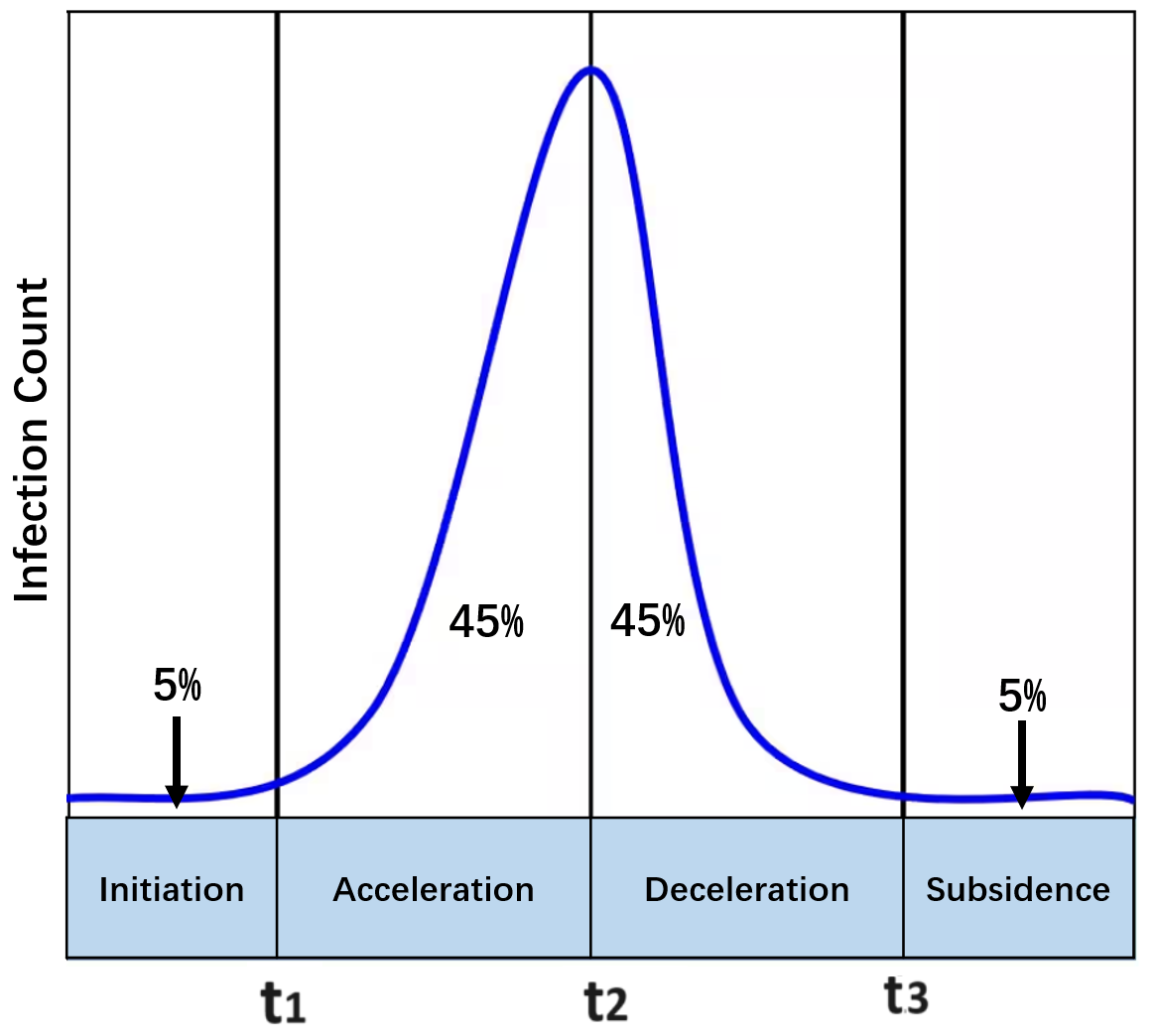}
  \vspace{-10pt}
  \caption{Pandemic Interval Division}
  \label{fig: pandemic_interval}
  \vspace{0pt}
\end{figure}

As shown in Fig.\ref{fig: pandemic_interval}The endpoints between these intervals are set based on the portion of total infected patient counts. If the duration of the total epidemic period is $T$ and it starts at $t_0$, the total infected population during the period is $\sum_{t_0}^T I(t)$. The endpoints $t_1$,$t_2$, and $t_3$ are defined as
\begin{equation*}
\centering
    \begin{cases}
        t_1: \sum_{t_0}^{t_1} I(t)/\sum_{t_0}^T I(t)=0.05,\\
        t_2: \sum_{t_0}^{t_2} I(t)/\sum_{t_0}^T I(t)=0.5,\\
        t_3: \sum_{t_0}^{t_3} I(t)/\sum_{t_0}^T I(t)=0.95.\\
    \end{cases}
\end{equation*}
\section{Bandwidth Selection for Kernel Density Estimation}\label{C: BS}
\begin{figure}[h]
\centering
\captionsetup{singlelinecheck = false, justification=justified}

  \includegraphics[scale=0.27]{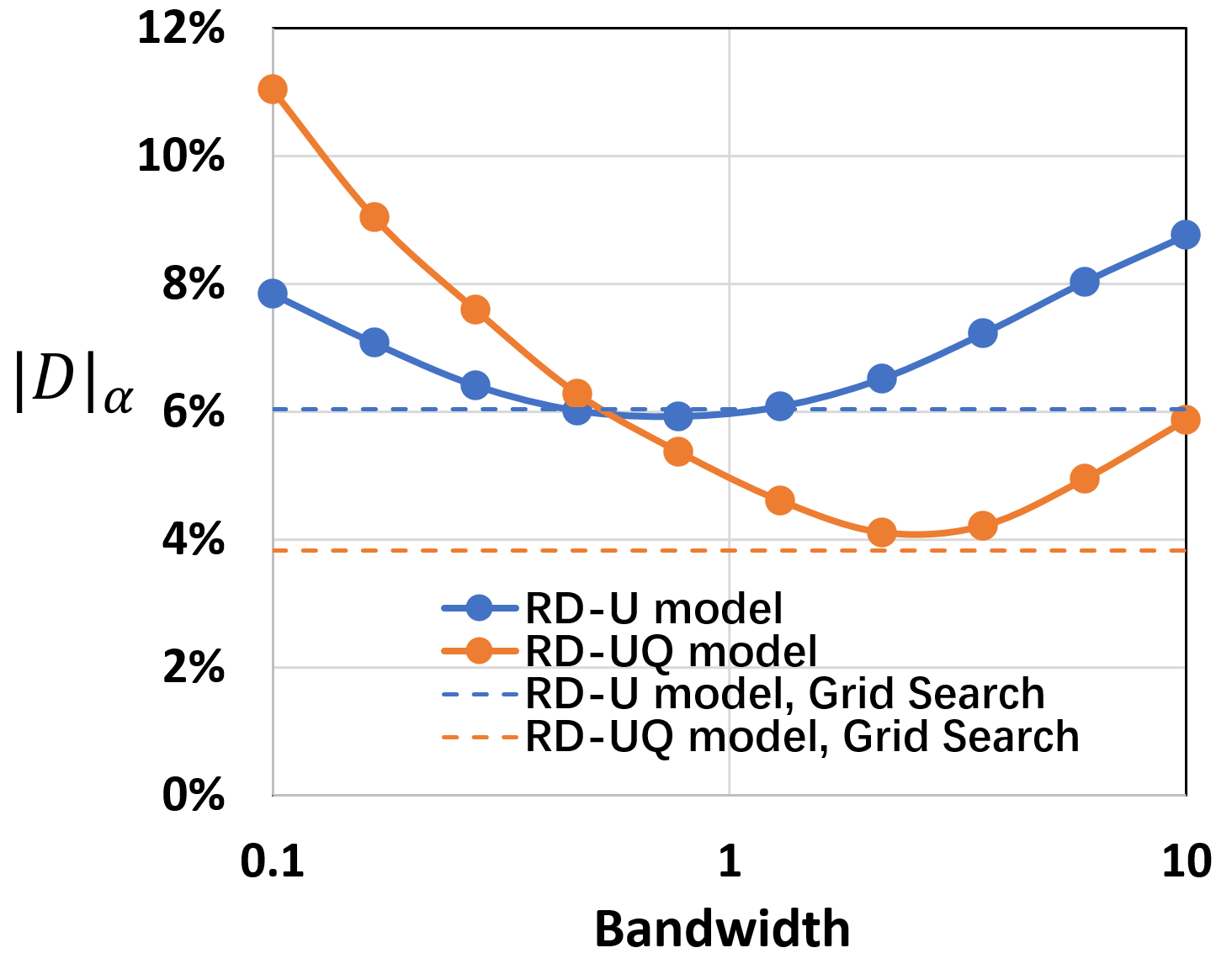}
  \vspace{-10pt}
  \caption{Illustration of KDE bandwidth impact on coverage divergence $|D|_\alpha$ for Seattle-loop dataset with $\alpha=0.2$. We can see the bandwidth optimized by grid search introduce lowest $|D|_\alpha$.}
  \label{fig: bandwidth_impact}
\end{figure}

For a fair comparison between models in terms of their coverage robustness, it is important to compare them with their own optimized hyperparameter values. Kernel density estimation (KDE) is applied when calculating the $w(X_i)=\frac{d\tilde P_X(X_i)}{dP_X(X_i)}$, which is influenced by the hyperparameter, bandwidth, $h$. In our experiment, we applied the Gaussian kernel, which is a positive function of a point $x$ as
\begin{equation*}
    K(x; h) \propto \exp(- \frac{x^2}{2h^2} ).
\end{equation*}
Given this kernel form, the density estimate at a position $x_p$ within a group of points $x_{i:n}$ is given by:
\begin{equation*}
    \rho_K(x_p) = \sum_{i=1}^{n} K(x_p - x_i; h).
\end{equation*}
To find the optimized bandwidth value for each model, we ran the experiment over a bandwidth pool and applied~\cite{scikit-learn} to check the outcomes from the grid search. We found the grid search can approximately obtain the minimum absolute coverage divergence of the results from the bandwidth pool, as Fig.\ref{fig: bandwidth_impact} shows, so we conduct the grid search to output a reliable experimental result.

\section{Experimental Results of Prediction Size}\label{D: PS}
\begin{figure}[h]
\centering
\captionsetup{singlelinecheck = false, justification=justified}

  \includegraphics[scale=0.27]{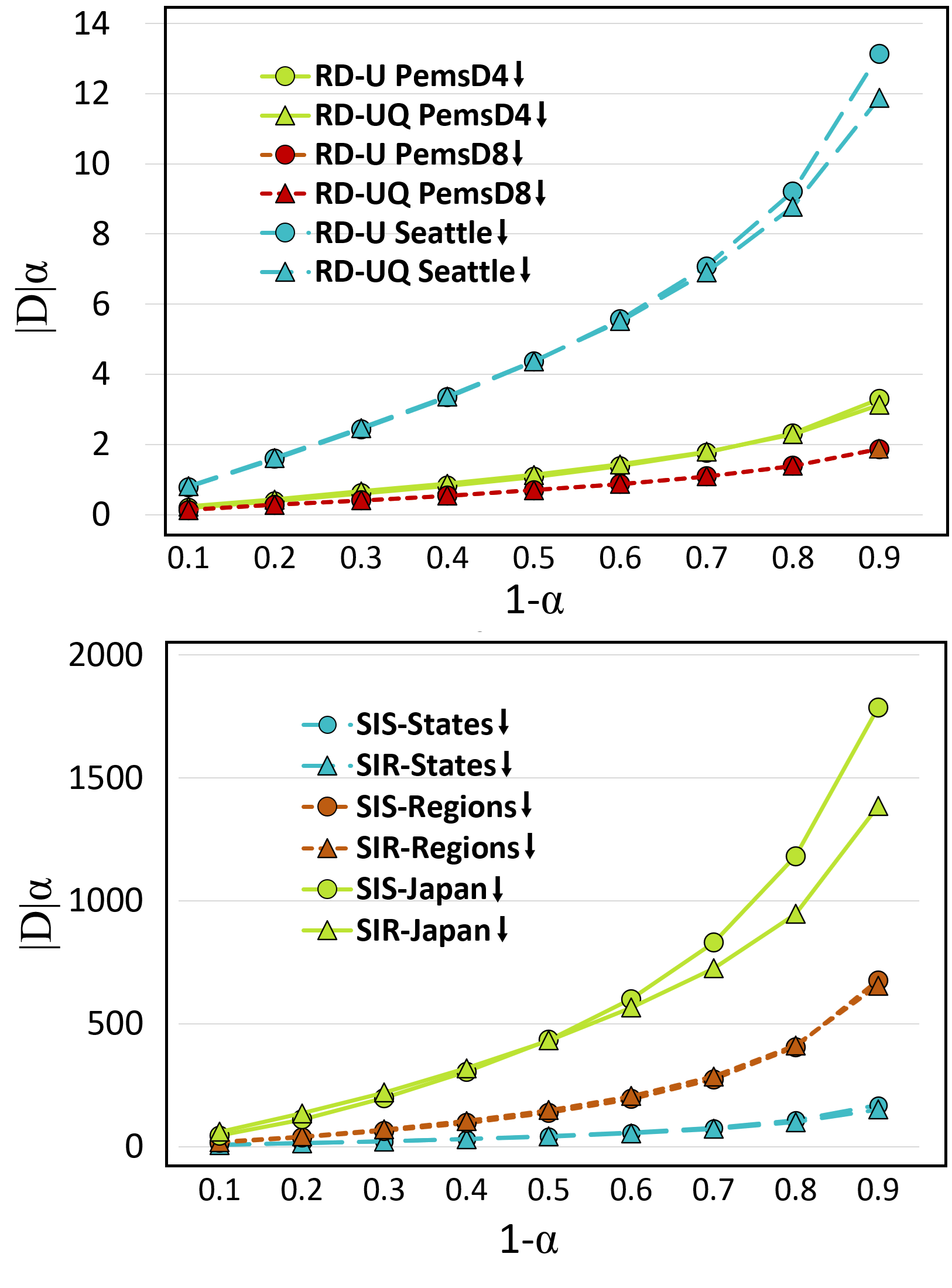}
  \vspace{-10pt}
  \caption{\textbf{Prediction Size along $(1-\alpha)$ confidence level of traffic speed prediction (top) and epidemic spread prediction (bottom)}.  The improvement in coverage robustness by PI-SCM does not increase prediction size as a sacrifice. In some cases, the prediction size by PI-SCM is even reduced. The result is averaged over 10 runs.}
  \label{fig: prediction_size}
\end{figure}

Prediction size is as important as coverage divergence in CP. Fig.\ref{fig: prediction_size} shows PI-SCM does not increase prediction size as a sacrifice of better coverage robustness. Prediction size can be influenced by multiple factors, like the variance and mean value of conformal score distribution. Usually, under the condition of exchangeability, a trained predictor with lower residuals will lead to a smaller prediction size. However, the situation will be more complicated under the non-exchangeable condition. If the coverage on test data by quantile $V_q$ is less than $(1-\alpha)$, it is inevitable to increase the prediction size for more accurate coverage. However, Fig.\ref{fig: prediction_size} shows PI-SCM can even reduce the prediction size in some experiment cases. We list possible two reasons here. First, the baseline model may be underconfident in its prediction sets. Secondly, PI-SCM reduces the mean and variance of weighted calibration conformal score distributions.


\end{document}